# Hybridization of Expectation-Maximization and K-Means Algorithms for Better Clustering Performance

*D. Raja Kishor*[*]*, N. B. Venkateswarlu*[**]

[*]*Ph.D. Student, Dept. of CSE, JNTU, Hyderabad, Telangana, India.*
[*]*Dept. of CSE, AITAM, Tekkali, Andhra Pradesh, India.*
*Emails: 1. rajakishor@gmail.com     2. venkat_ritch@yahoo.com*

**Abstract:** *The present work proposes hybridization of Expectation-Maximization (EM) and K-Means techniques as an attempt to speed-up the clustering process. Though both K-Means and EM techniques look into different areas, K-means can be viewed as an approximate way to obtain maximum likelihood estimates for the means. Along with the proposed algorithm for hybridization, the present work also experiments with the Standard EM algorithm. Six different datasets are used for the experiments of which three are synthetic datasets. Clustering fitness and Sum of Squared Errors (SSE) are computed for measuring the clustering performance. In all the experiments it is observed that the proposed algorithm for hybridization of EM and K-Means techniques is consistently taking less execution time with acceptable Clustering Fitness value and less SSE than the standard EM algorithm. It is also observed that the proposed algorithm is producing better clustering results than the Cluster package of Purdue University.*

## 1. Introduction

The Expectation Maximization (EM) algorithm is a model-based clustering technique, which attempts to optimize the fit between the given data and some mathematical model. Such methods are often based on the assumption that the data are generated by a mixture of underlying probability distributions [1].

The EM is an effective, popular technique for estimating mixture model parameters (like cluster weights and means) [7], [8], [9]. When compared to other clustering algorithms, the EM algorithm demands more computational efforts though it produces exceptionally good results [20], [21], [22]. Many researchers experimented on some variants (like generalized EM (GEM), Expectation Conditional Maximization (ECM), Sparse EM (SpEM), Lazy EM (LEM), Expectation-Conditional Maximization Either (ECME) algorithm and the Space Alternating Generalized EM (SAGE) algorithms) in order to reduce the execution time of EM algorithm [17], [18]. In [19], the use of Winograd's algorithm is proposed to reduce the computational efforts of E-step and M-step of the standard EM algorithm. In [15], the use of multi-criteria models is proposed to design clusters with the aim of improved clustering performance. All their experiments aimed at the speed-up of the EM algorithm by yielding the same results as the Standard EM algorithm or the better results without sacrificing its simplicity and stability.

As an attempt to speed-up the clustering process, the present work proposes the hybridization of EM and K-Means algorithms. The K-Means algorithm is a very popular algorithm for data clustering, which aims at the local minimum of the distortion [2], [23]. EM is a model based approach, which aims at finding clusters such that maximum likelihood of each cluster's parameters is obtained. In EM, each observation belongs to each cluster with a certain probability [2]. The K-means algorithm is the $2^{nd}$ dominantly used data mining algorithm and the EM algorithm is the $5^{th}$ dominantly used data mining algorithm [3], [4], [24]. Though both K-Means and EM techniques look into different areas [2], [23], K-means can be viewed as an approximate way to obtain maximum likelihood estimates for the means, which is the goal of density estimation in EM [23], [24].

In the present work, along with the proposed algorithm for hybridization of EM and K-Means techniques, experiments are carried out with the standard EM algorithm. In all the experiments, it is observed that the proposed algorithm for hybridization of EM and K-Means techniques is consistently taking less execution time to produce the clustering results with acceptable clustering fitness value and less SSE as compared to the standard EM algorithm. The proposed algorithm is also observed to produce clustering results with better performance than the Cluster Package of Purdue University [26].

## 2. The Standard EM (StEM) algorithm

The Expectation-Maximization (EM) algorithm partitions the given data by calculating the maximum a posteriori principle using the conditional probabilities [17]. Given a guess for the parameter values, the EM algorithm calculates the probability that each point belongs to each distribution and then uses these probabilities to compute a new estimate for the parameter. The EM algorithm iteratively refines initial mixture model parameter estimates to better fit the data and terminates at a locally optimal solution.

The standard EM [10], [11] for Gaussian Mixture Models (GMM) assumes that the algorithm will estimate k class distributions $C_j$, $j=1, \ldots, k$. For each of the input vectors $X_i$, $i=1, \ldots, N$, the algorithm calculates the probability $P(C_j/X_i)$. The highest probability will point to the vector's class.

The EM algorithm works iteratively by applying two steps: the Expectation step (E-step) and the Maximization step (M-step). Formally, $\hat{\theta}(t) = \{\mu_j(t), \Sigma_j(t), W_j(t)\}; j = 1,...,k$ stands for successive parameter estimates.

Given a dataset of $N$, $d$-dimensional vectors, the EM algorithm has to cluster them into $k$ groups.

The multi-dimensional Gaussian distribution for the cluster $C_j$ is parameterized by the $d$-dimensional mean column vector $\mu_j$ and $d \times d$ covariance matrix $\Sigma_j$ is given as follows [10]:

$$P(X_i | C_j) = \frac{1}{\sqrt{(2\Pi)^d |\Sigma_j|}} e^{-\frac{1}{2}(X_i-\mu_j)^T (\Sigma_j)^{-1}(X_i-\mu_j)} \quad (1)$$

where $X_i$ is a sample column vector, the superscript T indicates transpose of a column vector, $|\Sigma_j|$ is the determinant of $\Sigma_j$ and $(\Sigma_j)^{-1}$ is its matrix inverse of covariance matrix $\Sigma_j$.

The mixture model probability density function [10] is

$$P(X_i) = \sum_{l=1}^{k} W_l P(X_i | C_j) \quad (2)$$

where $W_l$ is the weight of cluster $C_l$.

## 2.1 Termination condition

As the termination condition, percentage change is computed using the following formula:

$$\text{Percentage change} = \frac{|\Psi_t - \Psi_{t+1}|}{\Psi_t} * 100 \quad (3)$$

where $\Psi_t$ is the number of vectors assigned to new clusters in $t^{th}$ iteration and $\Psi_{t+1}$ is the number of vectors assigned to new clusters in $(t+1)^{th}$ iteration. The symbol * indicates multiplication. The algorithm terminates when the *percentage change* < 3.

The EM algorithm for Gaussian Mixture Model [10] proceeds as follows:
1. Initialize mixture model parameters: set the current iteration $t=0$; set initial weights, $W$, to $1/k$ for all $k$ clusters; select $k$ vectors randomly from the dataset as the initial cluster means, $\mu$; compute global covariance matrix for the dataset and set it to be the initial covariance matrix, $\Sigma$, for all clusters.
2. E-step: Estimate the probability of each class $C_j$ ($j=1, 2, \ldots, k$), given a certain vector $X_i$ ($i=1, 2, \ldots, N$) for current iteration $t$ using the following formula and assign $X_i$ to the cluster with the maximum probability.

$$P(C_j | X_i) = \frac{W_j P(X_i | C_j)}{P(X_i)}$$

$$= \frac{|\Sigma_j(t)|^{-1/2} \exp^{\eta_j} . W_j(t)}{\sum_{l=1}^{k} |\Sigma_l(t)|^{-1/2} \exp^{\sigma_l} . W_l(t)} \quad (4)$$

where

$$\eta_j = -\frac{1}{2}(X_i - \mu_j(t))^T \Sigma_j^{-1}(t)(X_i - \mu_j(t))$$

$$\sigma_l = -\frac{1}{2}(X_i - \mu_l(t))^T \Sigma_l^{-1}(t)(X_i - \mu_l(t))$$

Each of the $k$ clusters has its mean ($\mu_j$) and covariance ($\Sigma_j$); $j = 1, 2, \ldots, k$. $W_j$ is the weight of $j^{th}$ cluster.

3. M-step: Here, for $j^{th}$ cluster, update the parameter estimation for the iteration $t+1$ as follows:

$$\mu_j(t+1) = \frac{\sum_{i=1}^{N} P(C_j | X_i) X_i}{\sum_{i=1}^{N} P(C_j | X_i)} \quad (5)$$

$$\sum_j(t+1) = \frac{\sum_{i=1}^{N} P(C_j \mid X_i)(X_i - \mu_j(t))(X_i - \mu_j(t))^T}{\sum_{i=1}^{N} P(C_j \mid X_i)} \quad (6)$$

$$W_t(t+1) = \frac{1}{N} \sum_{i=1}^{N} P(C_j \mid X_i) \quad (7)$$

4. Compute *percentage change* using (3).
5. Stop the process if the *percentage change* is < 3. Otherwise, set $t=t+1$ and repeat the steps 2 to 4 with the updated parameters.

## 3. Hybridization of EM and EM (HbEMKM) algorithms

Though an effectively used algorithm, the EM suffers from slow convergence as it requires heavy computational efforts involving in repeated computation of the many parameters like covariance matrices, means and weights of the clusters and repeated computation of the inverses of covariance matrices of the clusters [3], [5], [24], [25]. On the other hand, the K-Means algorithm can be used to simplify the computation and accelerate convergence as it requires only one parameter to compute, i.e., cluster means [23], [24]. While assigning points to the clusters, the EM maximizes the likelihood and the K-means minimizes the distortion with respect to the clusters [23].

The algorithm for conventional K-means is given below [12].

**Algorithm KMeans**
Select $k$ vectors randomly from the dataset as the initial cluster means, $\mu$. Set the current iteration $t=0$.
Repeat
    Assign each vector $X_i$ from the dataset to its closest cluster mean using Euclidean distance.

$$dist(X_i, \mu_j) = \sqrt{\sum_{l=1}^{d}(x_{il} - \mu_{lj})^2} \quad (8)$$

where $X_i$ is the $i^{th}$ vector in the dataset, $\mu_j$ is the mean of the cluster $j$ and $d$ is the number of dimensions of a data point.
    Re-compute the cluster means and set $t=t+1$.
    Compute *percentage change* using (3).
Until *percentage change* is < 3.
End of KMeans

The present work, as an attempt to speed-up the clustering process, experiments with the hybridization of EM and K-Means algorithms (HbEMKM). Though both EM and K-Means techniques look into different areas [2], [23], K-means can be viewed as an approximate way to obtain maximum likelihood estimates for the means, which is the goal of density estimation in EM [23], [24]. Furthermore, K-Means is formally equivalent to EM as K-Means is a limiting case of fitting data by a mixture of $k$ Gaussians with identical, isotropic covariance matrices ($\Sigma = \sigma^2 \mathbf{I}$), when the soft assignments of data points to mixture components are hardened to allocate each data point solely to the most likely component [3], [23]. A random space is isotropic if its covariance function depends on distance alone [25]. In practice, there is often some conflation of the two algorithms that K-means is sometimes used in density estimation applications due to its more rapid convergence [23].

Also that selection of initial values is critical for EM, since it most likely converges to local maxima around the initial values as EM uses maximum likelihood [2]. It may be a good practice, if the results of K-Means are used as initial parameter values for a subsequent execution of EM for the more exact computations [23], [24]. The present work also experiments on running the EM algorithm on the results of K-Means algorithm (*KMEM*).

Along with the proposed algorithm for hybridization of EM and K-Means techniques, experiments are carried out with the standard EM algorithm and finally performance comparison is made among the results of all experiments. In all the experiments same termination condition, discussed section 2.1, is used.

The pseudo code for the algorithm is given below. This algorithm performs clustering using Expectation-Maximization and K-Means techniques in the alternative iterations till termination. As part of maximization step for EM, cluster weights, means and covariance matrices are calculated using the results of K-Means step.

**Algorithm HbEMKM**
$N$ = number of samples in data
$n_j$ = number of samples in the $j^{th}$ cluster
$X_i$ = $i^{th}$ sample in data
$k$ = number of clusters
$W_j$ = weight of $j^{th}$ clusters
$\mu_j$ = mean of $j^{th}$ cluster
$\Sigma_j$ = covariance matrix of $j^{th}$ cluster
Select $k$ vectors randomly from the input dataset as the initial cluster means, $\mu$.
First, assign each data vector $X_i$ to the closest cluster with mean, $\mu_j$ using Euclidean distance in the formula (8).
Set *isProgress* = true
Repeat while (*isProgress* == true)
    M-Step: Compute means $\mu_j$ and covariance matrices $\Sigma_j$ for $j$ = 1, ..., $k$, based on the results of K-Means step.
    Compute cluster weights $W_j = n_j/N$ for $j$ = 1, ..., $k$.
    E-Step: For each given data vector $X_i$ ($i$ = 1, 2, ..., $N$), compute the cluster probability $P(C_j/X_i)$ for $j$ = 1, ..., $k$, using (4).
    Assign $X_i$ to the cluster with $\underset{j}{Max}\{P(C_j | X_i); j = 1,...,k\}$.
    Compute *percentage change* using (3).
    IF (*percentage change* >= 3)
        Compute cluster means $\mu_j$ for $j$ = 1, ..., $k$, using (5).
        K-Means Step: Assign each data vector $X_i$ to the closest cluster with mean, $\mu_j$ using Euclidean distance in the formula (8).
        Compute *percentage change* using (3).
        IF (*percentage change* >= 3)
            Set *isProgress* = true
        ELSE
            Set *isProgress* = false
        End of inner IF
    ELSE
        Set *isProgress* = false
    End of outer IF
End of Repeat Loop
End of HbEMKM

## 4. Clustering performance measure

As a measure of clustering performance, the *Clustering Fitness* [13] is computed. The calculation of *Clustering Fitness* involves intra-cluster similarity, inter-cluster similarity, and the experiential knowledge, $\lambda$. The main objective of any clustering algorithm is to generate clusters with higher intra-cluster similarity and lower inter-cluster similarity [16]. So both the measures are taken into consideration for computing *Clustering Fitness*. The computation of *Clustering Fitness* results in higher value when the inter-cluster similarity is low and results in lower value for when the inter-cluster similarity is high. To make the computation of *Clustering Fitness* unbiased, the value of $\lambda$ is taken as 0.5 [13].

4.1 Intracluster similarity for the cluster $C_j$

It can be quantified via some function of the reciprocals of intracluster radii within each of the resulting clusters. The intracluster similarity of a cluster $C_j$ (1 = j = k), denoted as $S_{tra}(C_j)$, is defined by

$$(9) \quad S_{tra}(C_j) = \frac{1+n_j}{1+\sum_1^{n_j} dist(I_l, Centroid)}$$

Here, $n_j$ is the number of items in cluster $C_j$, 1 = l = $n_j$, $I_l$ is the $l^{th}$ item in cluster $C_l$, and $dist(I_l, Centroid)$ calculates the distance between $I_l$ and the centroid of $C_j$, which is the intracluster radius of $C_j$. To smooth the value of $S_{tra}(C_j)$ and allow for possible singleton clusters 1 is added to the denominator and numerator.

4.2 Intracluster similarity for one clustering result C

Denoted as $S_{tra}(C)$, Intracluster similarity for one clustering result C is defined by

$$(10) \quad S_{tra}(c) = \frac{\sum_1^k S_{tra}(C_j)}{k}$$

Here, k is the number of resulting clusters in C.

4.3 Intercluster similarity

It can be quantified via some function of the reciprocals of intercluster radii of the clustering centroids. The intercluster similarity for one of the possible clustering results C, denoted as $S_{ter}(C)$, is defined by

$$(11) \quad S_{ter}(C) = \frac{1+n}{1+\sum_1^k dist(Centroid_j, Centroid^2)}$$

Here, k is the number of resulting clusters in C, 1 =j= k, $Centroid_j$ is the centroid of the $j^{th}$ cluster in C, $Centroid^2$ is the centroid of all centroids of clusters in C. We compute intercluster radius of $Centroid_j$ by calculating $dist(Centroid_j, Centroid^2)$, which is distance between $Centroid_j$, and $Centroid^2$. To smooth the value of $S_{ter}(C)$ and allow for possible all-inclusive clustering result, 1 is added to the denominator and the numerator.

4.4 Clustering fitness

The *Clustering Fitness* for one of the possible clustering results C, denoted as CF, is defined by

$$(12) \quad CF = \lambda * S_{tra} + \frac{1-\lambda}{S_{ter}}$$

Here, 0 < $\lambda$ < 1 is an experiential weight. The symbol * indicates multiplication. In the present work takes $\lambda$=0.5.

4.5 Sum of Squared Errors

In the present work, Sum of Squared Errors (SSE) is also computed for all the clustering results to measure the clustering performance [6]. The clustering performance is considered to be good if the corresponding SSE is less when compared to the other clustering techniques. The SSE is computed using the following formula.

$$(13) \quad SSE = \frac{1}{N} \sum_{j=1}^k \sum_{X_i \in C_j} |X_i - \mu_j|$$

Here, $X_i$ is a vector from the dataset, $\mu_j$ is the means of the cluster $C_j$, $k$ is the number of clusters and $N$ is the number of vectors in the dataset. $|X_i-\mu_j|$ denotes the distance between $X_i$ and $\mu_j$. The objective of clustering is to minimize the within-cluster sum of squared errors. The lesser the SSE, the better the goodness of fit is.

## 5. Experiment and results

Experiments are carried out on the system with Intel(R) Core i7-3770K with 3.50GHz processor speed, 8GB RAM with 1666FSB, Windows 7 OS and using JDK1.7.0_45. Separate modules are written for each algorithm to observe the CPU time for clustering any dataset by keeping the cluster seeds same for all methods. I/O operations are eliminated and time observed is strictly for clustering of the data.

Magic Gamma, Poker Hand, and Letter Recognition datasets are used for the present work from UCI ML dataset repository [14]. An important issue in evaluating data analysis algorithms is the availability of representative data. When real-life data are hard to obtain or when their properties are hard to modify for testing various algorithms, synthetic data becomes an appealing alternative. The present work also uses three synthetic datasets that are generated by an algorithm for generating multivariate normal random variables [27]. First synthetic dataset is generated assuming all clusters having different means and different covariance matrices. The second synthetic dataset is generated assuming some clusters having the same mean but different covariance matrices. The third synthetic dataset is generated assuming some clusters having same covariance matrix but different means.

| Data Set | No. of points | No. of Dimensions |
| --- | --- | --- |
| Letter Recognition Data | 20000 | 16 |
| Magic Gamma data | 19020 | 10 |
| Poker Hand data | 1025010 | 10 |
| Synthetic data-1 | 50000 | 10 |
| Synthetic data-2 | 50000 | 10 |
| Synthetic data-3 | 50000 | 10 |

All the algorithms are studied by executing on each dataset by varying number of clusters (i.e., k=10, 11, 12, 13, 14, 15). The details of execution time, clustering fitness and SSE of each algorithm are separately given in the tables below for each dataset.

5.1 Observations on Letter Recognition Dataset

The tables 1, 2 and 3 consist of the execution time, the cluster fitness and Sum of Squared Error (SSE), respectively, of algorithms discussed in sections 2 and 3 and the Cluster Package of Purdue University performed on Letter Recognition dataset. The observations are also shown in the Figures 1, 2 and 3.

Table 1. Execution time of each clustering method in seconds.

| K | StEM | KMEM | HbEMKM | Cluster 3.6.7 |
| --- | --- | --- | --- | --- |
| 10 | 16.4120 | 5.9210 | 0.2440 | 17.5760 |
| 11 | 6.4330 | 11.4470 | 0.3360 | 20.0250 |
| 12 | 14.0750 | 6.5540 | 0.4930 | 26.4340 |
| 13 | 7.6010 | 5.8790 | 0.3880 | 24.1890 |
| 14 | 5.1380 | 8.9930 | 0.4240 | 30.7420 |
| 15 | 13.5860 | 6.1920 | 0.7500 | 41.3680 |

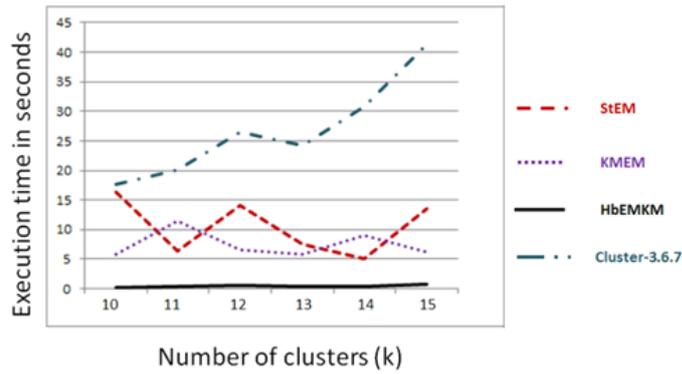

Fig. 1. Letter recognition dataset: Execution times

Table 2. Clustering fitness of each clustering method.

| K | StEM | KMEM | HbEMKM | Cluster 3.6.7 |
|---|------|------|--------|---------------|
| 10 | 2.6258 | 2.7286 | 2.9425 | 2.7812 |
| 11 | 2.6063 | 2.9516 | 3.1282 | 2.7391 |
| 12 | 2.7610 | 2.8908 | 3.1003 | 2.7088 |
| 13 | 2.8867 | 3.1064 | 3.2270 | 2.9507 |
| 14 | 3.0719 | 3.3224 | 3.4473 | 2.8795 |
| 15 | 2.9324 | 3.0599 | 3.3460 | 3.0419 |

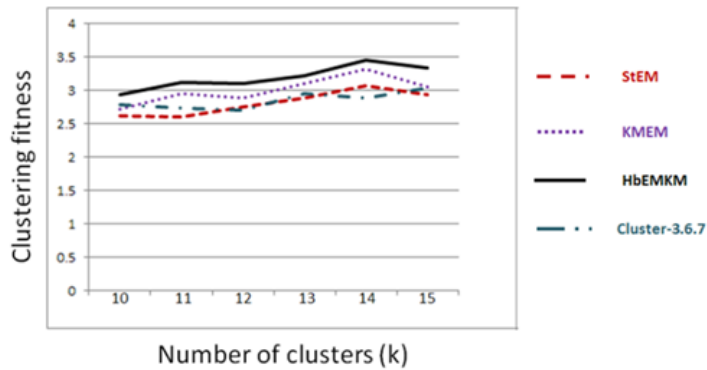

Fig. 2. Letter recognition dataset: Clustering fitness

Table-3: SSE of each clustering method.

| K | StEM | KMEM | HbEMKM | Cluster 3.6.7 |
|---|------|------|--------|---------------|
| 10 | 58.31780253 | 54.90029903 | 46.79034965 | 59.54343871 |
| 11 | 56.81365607 | 50.53051120 | 43.76738443 | 59.06663898 |
| 12 | 55.26046620 | 49.59847946 | 43.60821184 | 55.55461758 |
| 13 | 52.24618292 | 47.06660999 | 41.57619654 | 54.98634779 |
| 14 | 52.43310568 | 46.24255323 | 39.85697243 | 53.61458606 |
| 15 | 52.74977604 | 47.52093477 | 39.76810321 | 51.85401078 |

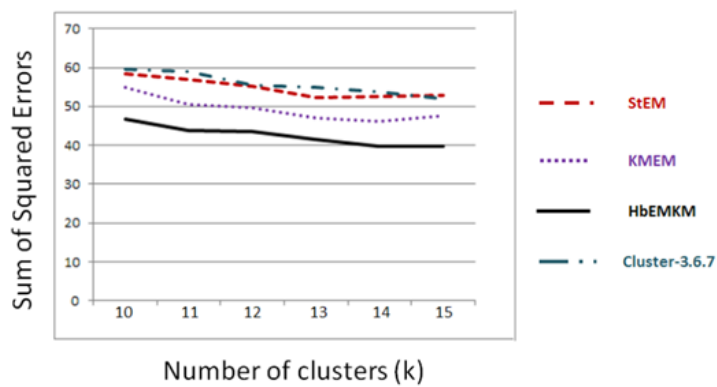

Fig. 3. Letter recognition dataset: Sum of squared errors

## 5.2 Observations on Magic Gamma dataset

The tables 4, 5 and consist of the execution time, the cluster fitness and Sum of Squared Error (SSE), respectively, of algorithms discussed in sections 2 and 3 and the Cluster Package of Purdue University performed on Magic Gamma dataset. The observations are also shown in the Figures 4, 5 and 6.

Table-4: Execution time of each clustering method in seconds.

| K | StEM | KMEM | HbEMKM | Cluster 3.6.7 |
|---|------|------|--------|---------------|
| 10 | 3.5360 | 0.7920 | 0.1830 | 7.2890 |
| 11 | 3.7360 | 4.0920 | 0.1110 | 10.2680 |
| 12 | 3.2410 | 3.1720 | 0.2420 | 9.3120 |
| 13 | 3.2390 | 4.9610 | 0.2570 | 11.0620 |
| 14 | 6.8540 | 7.9350 | 0.4070 | 9.5590 |
| 15 | 5.9050 | 3.3490 | 0.5020 | 18.6780 |

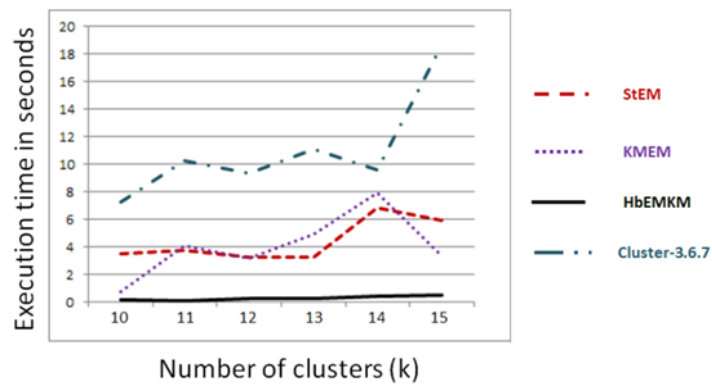

Fig. 4. Magic gamma dataset: Execution times

Table-5: Clustering fitness of each clustering method.

| K | StEM | KMEM | HbEMKM | Cluster 3.6.7 |
|---|------|------|--------|---------------|
| 10 | 29.8434 | 46.1219 | 50.8960 | 34.9529 |
| 11 | 35.9602 | 40.2215 | 46.6300 | 36.0443 |
| 12 | 37.5347 | 44.2417 | 57.5562 | 38.4638 |
| 13 | 34.8555 | 41.3342 | 52.9504 | 40.7634 |
| 14 | 33.7390 | 39.9490 | 52.9743 | 41.0526 |
| 15 | 42.8234 | 46.1663 | 61.7020 | 39.9385 |

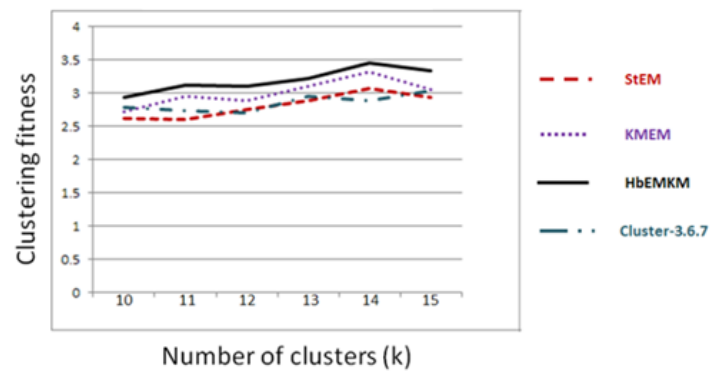

Fig. 5. Magic gamma dataset: Clustering fitness

Table-6: SSE of each clustering method.

| K | StEM | KMEM | HbEMKM | Cluster 3.6.7 |
|---|------|------|--------|---------------|
| 10 | 10235.04311 | 6336.868507 | 5116.794358 | 9619.032540 |
| 11 | 9578.544208 | 8140.627114 | 5013.559560 | 8606.290689 |
| 12 | 9873.706092 | 7544.572959 | 4409.876943 | 8838.627792 |
| 13 | 9272.422300 | 8045.302214 | 4658.203266 | 8743.184190 |
| 14 | 9398.092913 | 7830.483924 | 4397.280683 | 8191.174528 |
| 15 | 8646.400371 | 6528.632337 | 3803.336041 | 8620.268594 |

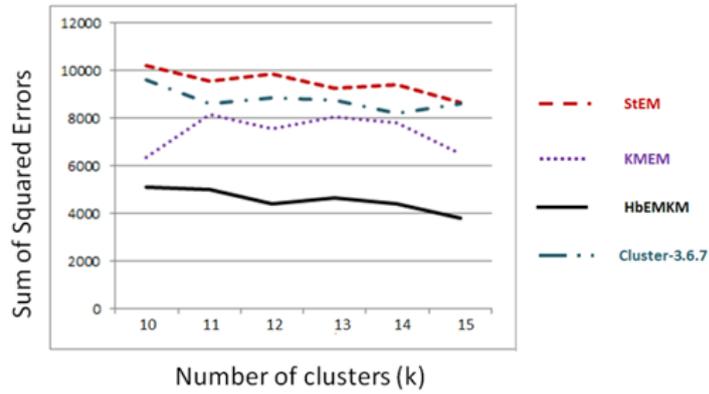

Fig. 6. Magic gamma dataset: Sum of squared errors

5.3 Observations on Poker hand dataset

The tables 7, 8 and 9 consist of the execution time, the cluster fitness and Sum of Squared Error (SSE), respectively, of algorithms discussed in sections 2 and 3 and the Cluster Package of Purdue University performed on Poker Hand dataset. The observations are also shown in the Figures 7, 8 and 9.

Table-7: Execution time of each clustering method in seconds.

| K | StEM | KMEM | HbEMKM | Cluster 3.6.7 |
|---|---|---|---|---|
| 10 | 66.1490 | 224.7950 | 15.5420 | 1016.9140 |
| 11 | 74.0110 | 107.9470 | 23.3930 | 2557.5940 |
| 12 | 78.4610 | 82.4270 | 31.0430 | 3328.3360 |
| 13 | 85.2930 | 72.0570 | 29.0900 | 3434.1700 |
| 14 | 91.0140 | 332.3750 | 46.2370 | 3160.3870 |
| 15 | 117.2070 | 238.8700 | 28.6360 | 2809.5350 |

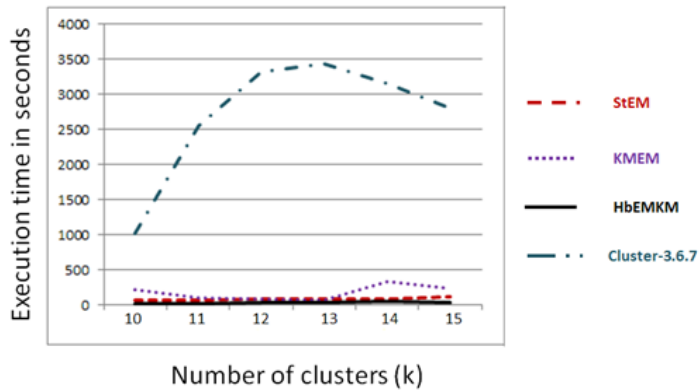

Fig. 7. Poker hand dataset: Execution times

Table-8: Clustering fitness of each clustering method.

| K | StEM | KMEM | HbEMKM | Cluster 3.6.7 |
|---|---|---|---|---|
| 10 | 1.3840 | 2.7620 | 2.8951 | 1.2293 |
| 11 | 1.6828 | 2.8351 | 2.9882 | 1.8606 |
| 12 | 1.5570 | 2.9154 | 3.0631 | 1.3815 |
| 13 | 1.4044 | 2.9795 | 3.1120 | 1.0973 |
| 14 | 1.5571 | 3.0186 | 3.1739 | 1.5624 |
| 15 | 1.8413 | 3.0663 | 3.2148 | 1.0539 |

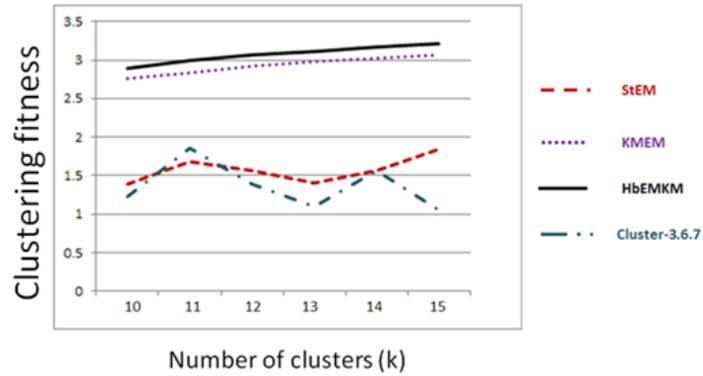

Fig. 8. Poker hand dataset: Clustering fitness

Table-9: SSE of each clustering method.

| K | StEM | KMEM | HbEMKM | Cluster 3.6.7 |
|---|---|---|---|---|
| 10 | 61.89049571 | 40.41746868 | 39.33511213 | 72.07099026 |
| 11 | 57.49226138 | 38.17264347 | 37.30469352 | 62.23416768 |
| 12 | 59.57113812 | 36.38456914 | 35.82492696 | 70.92287882 |
| 13 | 61.55035153 | 35.23419731 | 34.83422455 | 72.85375279 |
| 14 | 58.56628704 | 34.80452906 | 33.53945731 | 73.87162564 |
| 15 | 56.33032095 | 33.82117461 | 32.75190094 | 71.93356843 |

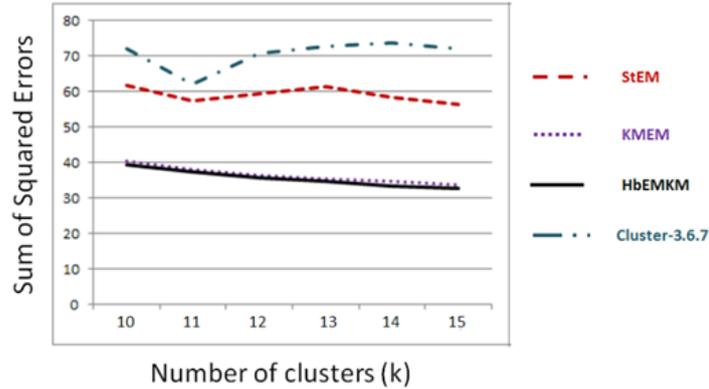

Fig. 9. Poker hand dataset: Sum of squared errors

5.4 Observations on Synthetic dataset-1

The tables 10, 11 and 12 consist of the execution time, the cluster fitness and Sum of Squared Error (SSE), respectively, of algorithms discussed in sections 2 and 3 and the Cluster Package of Purdue University performed on Synthetic dataset-1. The observations are also shown in the Figures 10, 11 and 12.

Table-10: Execution time of each clustering method in seconds.

| K | StEM | KMEM | HbEMKM | Cluster 3.6.7 |
|---|---|---|---|---|
| 10 | 3.8230 | 3.3670 | 0.6580 | 43.3850 |
| 11 | 5.6100 | 2.9100 | 0.8740 | 46.5140 |
| 12 | 6.1160 | 3.9880 | 0.9400 | 59.1600 |
| 13 | 4.9520 | 4.2760 | 1.3410 | 70.7840 |
| 14 | 4.4550 | 3.7160 | 1.0860 | 87.9630 |
| 15 | 4.7690 | 6.8270 | 0.9680 | 106.1600 |

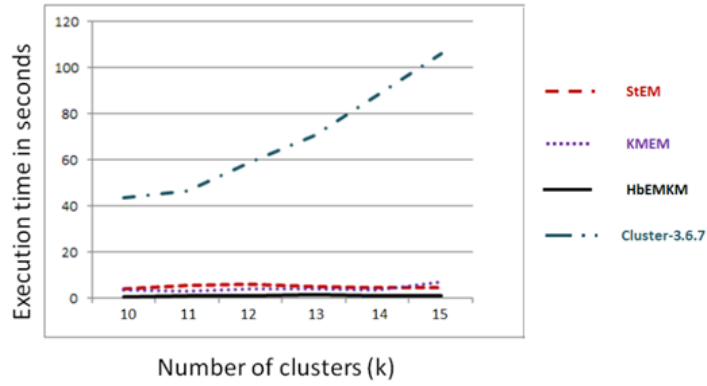

Fig. 10. Synthetic dataset-1: Execution times

Table-11: Clustering fitness of each clustering method.

| K | StEM | KMEM | HbEMKM | Cluster 3.6.7 |
|---|------|------|--------|---------------|
| 10 | 582.8037 | 989.8852 | 1070.4906 | 862.1022 |
| 11 | 719.8391 | 1002.2908 | 1080.2918 | 901.5273 |
| 12 | 690.8997 | 1043.5350 | 1135.5673 | 897.9433 |
| 13 | 663.8376 | 1049.8692 | 1143.5548 | 380.6306 |
| 14 | 763.8713 | 1081.2933 | 1165.1284 | 443.3503 |
| 15 | 842.4782 | 1107.4315 | 1200.3491 | 361.4750 |

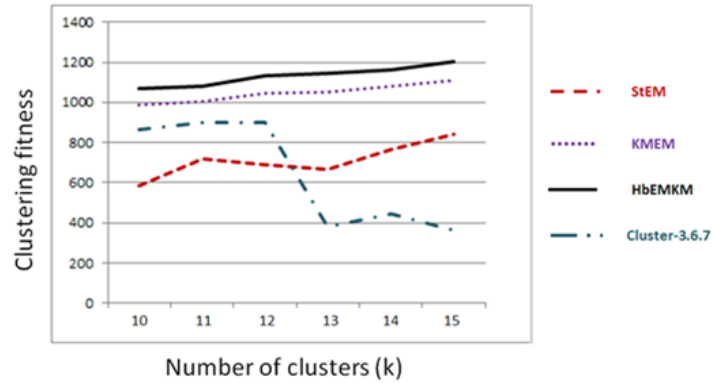

Fig. 11. Synthetic dataset-1: Clustering fitness

Table-12: SSE of each clustering method.

| K | StEM | KMEM | HbEMKM | Cluster 3.6.7 |
|---|------|------|--------|---------------|
| 10 | 14628710.78 | 11859786.85 | 11684878.89 | 14388425.51 |
| 11 | 14386916.86 | 11658179.86 | 11468010.80 | 13455719.67 |
| 12 | 13874673.81 | 11491360.59 | 11251086.05 | 13535376.15 |
| 13 | 14099367.22 | 11385248.64 | 11110137.28 | 20481277.55 |
| 14 | 13589123.08 | 11212795.22 | 11046519.80 | 25434272.93 |
| 15 | 13452374.50 | 11267402.97 | 10862782.33 | 21592712.96 |

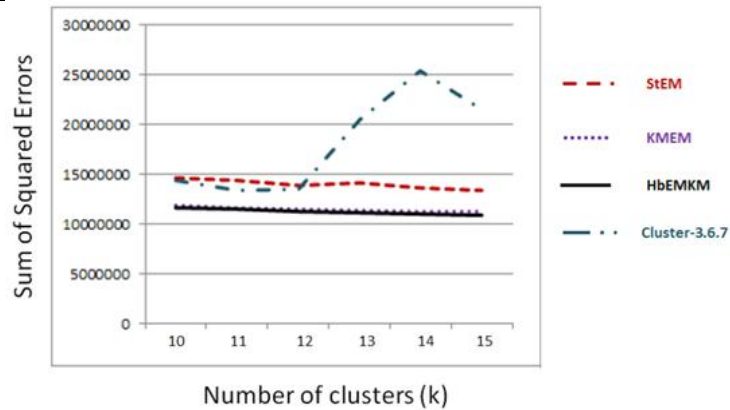

Fig. 12. Synthetic dataset-1: Sum of squared errors

## 5.5 Observations on Synthetic dataset-2

The tables 13, 14 and 15 consist of the execution time, the cluster fitness and Sum of Squared Error (SSE), respectively, of algorithms discussed in sections 2 and 3 and the Cluster Package of Purdue University performed on Synthetic dataset-2. The observations are also shown in the Figures 13, 14 and 15.

Table-13: Execution time of each clustering method in seconds.

| K | StEM | KMEM | HbEMKM | Cluster 3.6.7 |
|---|------|------|--------|---------------|
| 10 | 3.8420 | 2.6790 | 0.6660 | 30.8250 |
| 11 | 4.9020 | 2.9210 | 0.8630 | 51.8800 |
| 12 | 4.5760 | 3.9120 | 0.9290 | 49.4800 |
| 13 | 7.4250 | 5.1100 | 1.0130 | 64.8300 |
| 14 | 7.1170 | 3.6620 | 1.0750 | 87.2220 |
| 15 | 5.7130 | 3.9850 | 1.1530 | 88.0840 |

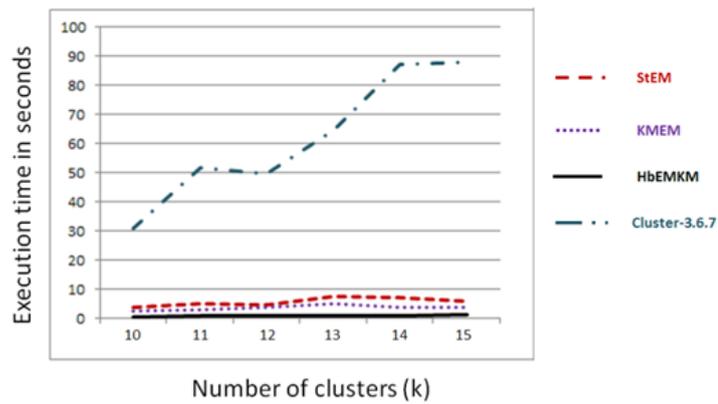

Fig. 13. Synthetic dataset-2: Execution times

Table-14: Clustering fitness of each clustering method.

| K | StEM | KMEM | HbEMKM | Cluster 3.6.7 |
|---|------|------|--------|---------------|
| 10 | 592.9165 | 985.5797 | 1064.7669 | 479.4660 |
| 11 | 705.1030 | 998.8906 | 1083.0545 | 505.4192 |
| 12 | 616.7459 | 1028.0205 | 1071.5066 | 391.3777 |
| 13 | 865.7343 | 1040.6712 | 1148.6615 | 435.4822 |
| 14 | 761.9546 | 1080.9615 | 1174.8407 | 321.3893 |
| 15 | 775.9452 | 1096.7161 | 1180.8662 | 412.5189 |

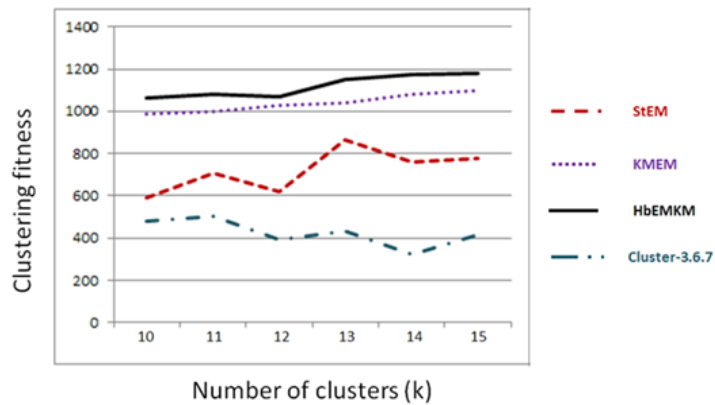

Fig. 14. Synthetic dataset-2: Clustering fitness

Table-15: SSE of each clustering method.

| K | StEM | KMEM | HbEMKM | Cluster 3.6.7 |
|---|------|------|--------|---------------|
| 10 | 14439664.00 | 11760241.27 | 11598373.61 | 20164991.50 |
| 11 | 13902439.10 | 11528187.47 | 11355916.26 | 21276089.23 |
| 12 | 14232153.20 | 11421357.02 | 11287930.24 | 21633831.68 |
| 13 | 13415780.69 | 11319653.67 | 11024847.95 | 21181605.38 |
| 14 | 13336198.27 | 11100700.03 | 10882717.53 | 23815685.30 |
| 15 | 13320932.07 | 10941351.39 | 10743553.26 | 21087020.41 |

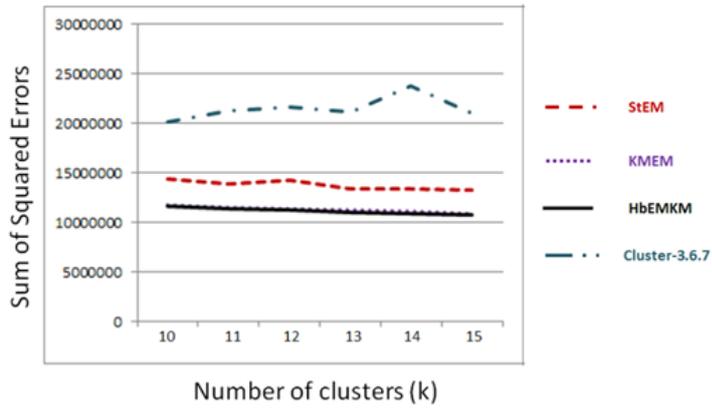

Fig. 15. Synthetic dataset-2: Sum of squared errors

5.6 Observations on Synthetic dataset-3

The tables 16, 17 and 18 consist of the execution time, the cluster fitness and Sum of Squared Error (SSE), respectively, of algorithms discussed in sections 2 and 3 and the Cluster Package of Purdue University performed on Synthetic dataset-3. The observations are also shown in the Figures 16, 17 and 18.

Table-16: Execution time of each clustering method in seconds.

| K | StEM | KMEM | HbEMKM | Cluster 3.6.7 |
|---|------|------|--------|---------------|
| 10 | 8.2650 | 2.0130 | 0.8030 | 31.9960 |
| 11 | 4.1960 | 3.6610 | 0.7170 | 60.8990 |
| 12 | 4.5890 | 2.4560 | 0.7870 | 48.1750 |
| 13 | 5.7860 | 4.2670 | 1.0130 | 70.2460 |
| 14 | 6.2350 | 8.0820 | 0.7250 | 62.2360 |
| 15 | 5.7190 | 4.9670 | 0.9640 | 95.2090 |

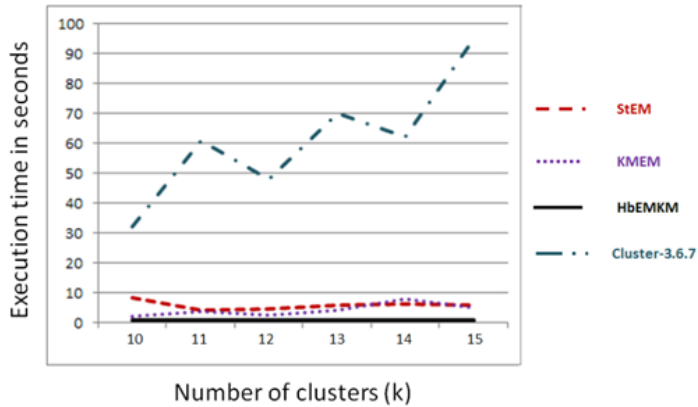

Fig. 16. Synthetic dataset-3: Execution times

Table-17: Clustering fitness of each clustering method.

| K | StEM | KMEM | HbEMKM | Cluster 3.6.7 |
|---|------|------|--------|---------------|
| 10 | 623.6250 | 972.4719 | 1048.2655 | 500.3920 |
| 11 | 671.4153 | 982.1240 | 1066.9712 | 391.8017 |
| 12 | 670.1226 | 1026.8664 | 1106.1198 | 419.4653 |
| 13 | 797.2359 | 1039.7200 | 1116.5491 | 372.0694 |
| 14 | 872.9962 | 1048.6642 | 1157.4691 | 397.6201 |
| 15 | 685.2117 | 1075.6261 | 1149.7448 | 332.1400 |

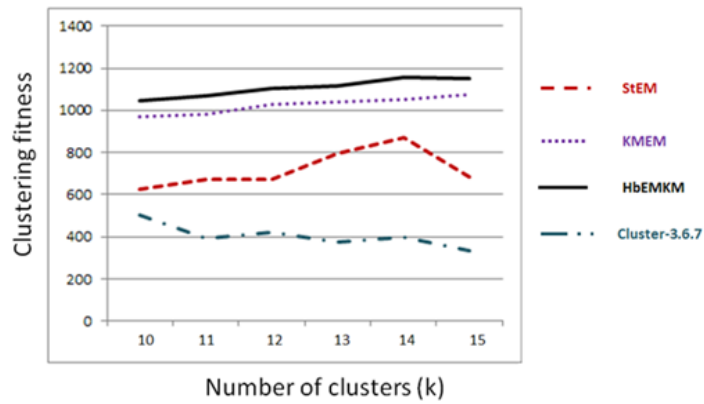

Fig. 17. Synthetic dataset-3: Clustering fitness

Table-18: SSE of each clustering method.

| K | StEM | KMEM | HbEMKM | Cluster 3.6.7 |
|---|---|---|---|---|
| 10 | 13456811.43 | 11348281.15 | 11197451.26 | 19005111.06 |
| 11 | 13810971.74 | 11258784.69 | 11072963.24 | 19915617.9 |
| 12 | 13190851.64 | 10946003.08 | 10842930.4 | 19915392.35 |
| 13 | 13235505.84 | 10976305.12 | 10717404.62 | 21071287.76 |
| 14 | 12527770.20 | 11236068.19 | 10616547.71 | 19920287.74 |
| 15 | 13107521.88 | 10636008.11 | 10406779.38 | 21017062.74 |

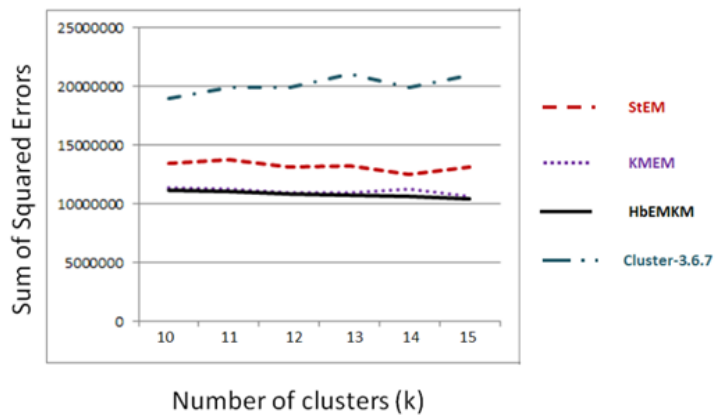

Fig. 18. Synthetic dataset-3: Sum of squared errors

## 6. Conclusion

The proposed algorithm for Hybridization of EM and K-Means is consistently taking less computational time with all the tested datasets. The algorithm also takes less computational time when compared to the Cluster-3.6.7 package of Purdue University. The proposed algorithm also produces the results with higher clustering fitness values than the other algorithms including Cluster-3.6.7. It is also observed that the proposed algorithm produces the clustering results with lesser SSE values than the other algorithms including the Cluster-3.6.7 package. Therefore, the present work proposes Hybridization of EM and K-Means algorithms as a faster clustering technique with improved performance.